\documentclass[10pt,journal,compsoc,final]{IEEEtran}
\usepackage{graphicx}
\usepackage[cmex10]{amsmath}
\usepackage{cite}
\usepackage{hyperref}

\usepackage{fixltx2e}
\usepackage{url}
\usepackage{amsfonts}

\usepackage{array}
\newcolumntype{P}[1]{>{\centering\arraybackslash}p{#1}}
\newcolumntype{M}[1]{>{\centering\arraybackslash}m{#1}}

\DeclareMathOperator*{\argmin}{arg\,min}
\newcommand{\norm}[1]{\left\lVert#1\right\rVert}
\newcommand{\supp}{\operatorname{supp}} 

\begin{document}

\title{Light Field Reconstruction\\Using Shearlet Transform}

\author{Suren~Vagharshakyan,
        Robert~Bregovic
        and~Atanas~Gotchev,~\IEEEmembership{Member,~IEEE}}

\IEEEtitleabstractindextext{
\begin{abstract}
In this article we develop an image based rendering technique based on light field reconstruction from a limited set of perspective views acquired by cameras. Our approach utilizes sparse representation of epipolar-plane images in a directionally sensitive transform domain, obtained by an adapted discrete shearlet transform. The used iterative thresholding algorithm provides high-quality reconstruction results for relatively big disparities between neighboring views. The generated densely sampled light field of a given 3D scene is thus suitable for all applications which requires light field reconstruction. The proposed algorithm is compared favorably against state of the art depth image based rendering techniques.
\end{abstract}

}

\maketitle
\IEEEdisplaynontitleabstractindextext
\IEEEpeerreviewmaketitle

\ifCLASSOPTIONcompsoc
\IEEEraisesectionheading{\section{Introduction}\label{sec:introduction}}
\else
\section{Introduction}
\label{sec:introduction}
\fi

\IEEEPARstart{S}{ynthesis} of intermediate views from a given set of captured views of a 3D visual scene is usually referred to as image-based rendering (IBR)~\cite{shum2008image}. The scene is typically captured by a limited number of cameras which form a rather coarse set of multiview images. However, denser set of images (i.e. intermediate views) is required in immersive visual applications such as free viewpoint television (FVT) and virtual reality (VR) aimed at creating the perception of continuous parallax.

Modern view synthesis methods are based on two, fundamentally different, approaches. The first approach is based on the estimation of the scene depth and synthesis of novel views based on the estimated depth and the given images, where the depth information works as correspondence map for view reprojection. A number of depth estimation methods have been developed specifically for stereo images~\cite{scharstein2002taxonomy}, and for multiview images as well ~\cite{kim2013scene}, ~\cite{pearson2013plenoptic}, ~\cite{wanner2014variational}. In all cases, the quality of depth estimation is very much content (scene) dependent. This is a substantial problem since small deviations in the estimated depth map might introduce visually annoying artifacts in the rendered (synthesized) views. The second approach is based on the concept of plenoptic function and its light field (LF) approximation ~\cite{adelson1991plenoptic}, ~\cite{levoy1996light}. The scene capture and intermediate view synthesis problem can be formulated as sampling and consecutive reconstruction (interpolation) of the underlying plenoptic function. LF based methods do not use the depth information as an auxiliary mapping. Instead, they consider each pixel of the given views as a sample of a multidimensional LF function, thus the unknown views are function values that can be determined after its reconstruction from samples. In ~\cite{gortler1996lumigraph}, different interpolation kernels utilizing available geometrical information are discussed. As shown there, established interpolation algorithms such as  linear interpolation require a substantial number of samples (images) in order to obtain synthesized views with good quality.

The required bounds for sampling the LF of a scene have been defined in ~\cite{lin2004geometric}. In order to generate novel views without ghosting effects by using linear interpolation, one needs to sample the LF such that the disparity between neighboring views is less than one pixel ~\cite{lin2004geometric}. Hereafter, we will refer to such sampling as dense sampling and to the correspondingly sampled LF as densely sampled LF. In order to capture a densely sampled LF, the required distance between neighboring camera positions can be estimated based on the minimal scene depth $(z_{min})$ and the camera resolution. Furthermore, camera resolution should provide enough samples to properly capture highest spatial texture frequency in the scene ~\cite{chai2000plenoptic}.

Densely sampled LF is an attractive representation of scene visual content, particularly for applications, such as refocused image generation~\cite{ng2005fourier}, dense depth estimation~\cite{tosic2014light}, novel view generation for FVT~\cite{tanimoto2006overview}, and holographic stereography~\cite{jurik2012geometry}. However, in many practical cases one is not able to sample a real-world scene with sufficient number of cameras to directly obtain a densely sampled LF. Therefore, the required number of views has to be generated from the given sparse set of images by using IBR.
The work~\cite{chai2000plenoptic} has discussed the effective use of the depth limits $(z_{min},z_{max})$ in order to reconstruct desired views from a limited number of given views using appropriate interpolation filters. Use has been made of the so-called epipolar-plane image (EPI) and its Fourier domain properties~\cite{bolles1987epipolar}. Further benefits in terms of improved rendering quality has been achieved by using depth layering ~\cite{pearson2013plenoptic}, ~\cite{chai2000plenoptic}. More recently, another approach to LF reconstruction has been proposed~\cite{shi2014light}. It considers the LF sampled by a small number of 1D viewpoint trajectories and employs sparsity in continuous Fourier domain in order to reconstruct the remaining full-parallax views. 

\begin{figure*}[t]
\centering
\includegraphics[width=6.5in]{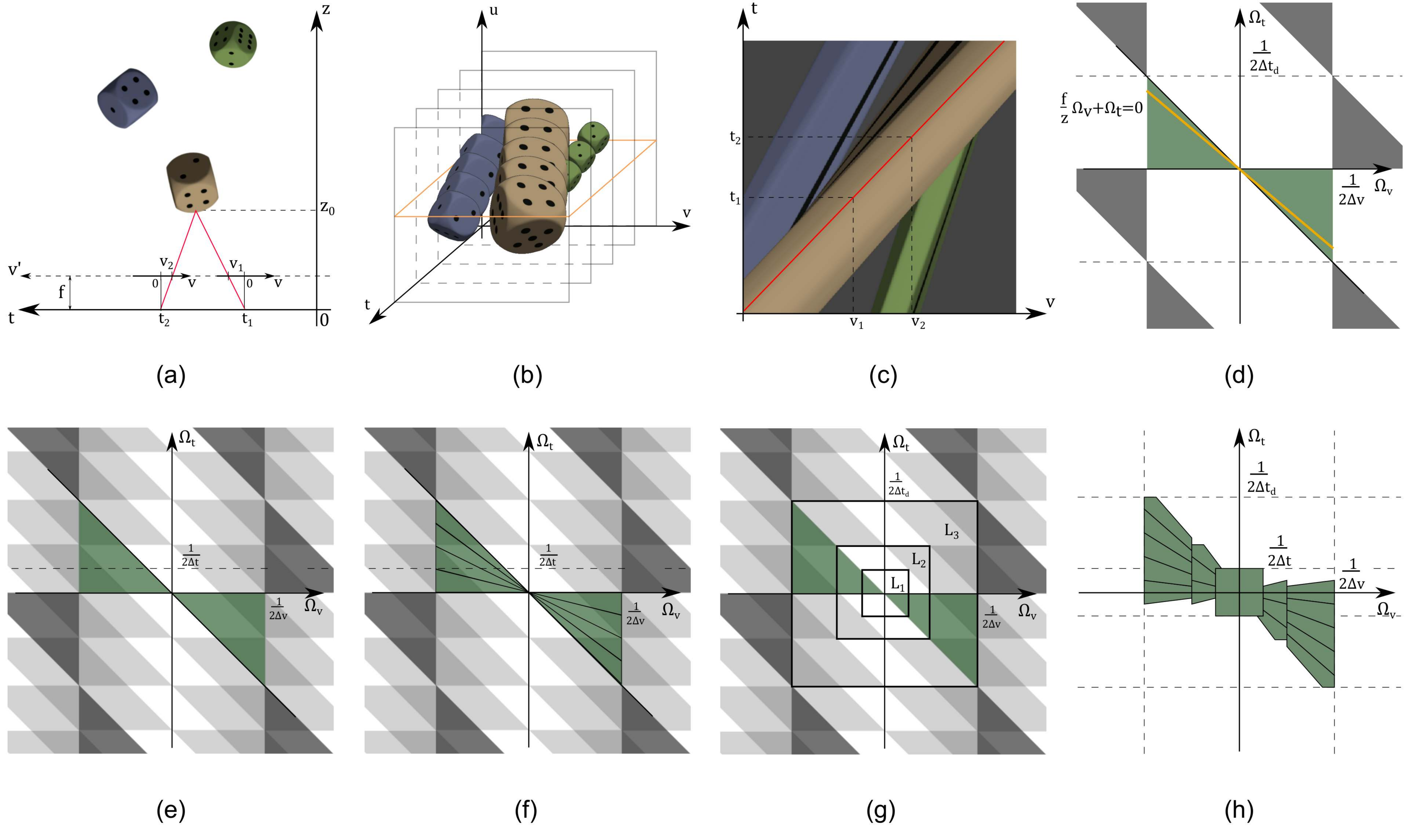}
\caption{Epipolar-plane image (EPI) formation and its frequency domain properties. (a) Capturing setup and EPI formation, a scene point is observed by two pinhole cameras positioned at $t_1$, $t_2$ at image coordinates $v_1$ and $v_2$ respectively; (b) Stack of captured images; an epipolar plane is highlighted for fixed vertical image coordinate $u$; (c) Example of EPI; red line represents a scene point in different cameras; (d) Frequency support of a densely sampled EPI; green are a represents the baseband bounded by min and max depth; yellow line corresponds to a depth layer, the slope determines the depth value; (e) Frequency domain structure of an EPI being insufficiently sampled over $t$-axis, the overlapping regions represent aliasing; (f) Desirable frequency domain separation based on depth layering; (g) Frequency domain separation based on dyadic scaling; (h) Composite directional and scaling based frequency domain separation for EPI sparse representation.}
\label{fig:epi}
\end{figure*}

In this article, we advance the concepts of LF sparsification and depth layering with the aim to develop an effective reconstruction of the LF represented by EPIs. The reconstruction utilizes the fact that EPIs have sparse representation in an appropriate transform domain. Furthermore, we also assume that a good sparse transform should incorporate scene representation with depth layers, which are expected to be sparse. We favor the shearlet transform as the sought sparsifying transform and develop an inpainting technique working on EPI, in a fashion similar to how shearlets have been applied for seismic data reconstruction ~\cite{hauser2012seismic}.

Preliminary results of novel view synthesis by using shearlet transform have been presented in~\cite{vagh2015imag}. In this paper, we extend the ideas presented in~\cite{vagh2015imag} by including the underlying analysis, describing in detail the construction of the used shearlet transform and the corresponding view synthesis algorithm and evaluating the efficiency of the proposed algorithm on various datasets.

The outline of this paper is as follows. The LF and EPI concepts are presented in Section~\ref{sec:lf}. Section~\ref{sec:epitd} focuses on the identification of suitable transform domain and particularly discusses the shearlet transform, its properties and construction for the given case. The reconstruction algorithm is presented in Section~\ref{sec:recalg}. The algorithm evaluation for different datasets and a comparison with the state of the art is presented in Section~\ref{sec:eval}. Finally, the work is concluded in Section~\ref{sec:conc}.

\section{Light field formalization}\label{sec:lf}

\subsection{Light field representation}
The propagation of light in space in terms of rays is fully described by the $7D$ continuous plenoptic function $R(\theta,\phi,\lambda,\tau,V_x,V_y,V_z)$, where $(V_x,V_y,V_z)$ is a location in the 3D space, $(\theta,\phi)$ are propagation angles, $\lambda$ is wavelength, and $\tau$ is time ~\cite{adelson1991plenoptic}. In more practical considerations, the plenoptic function is simplified to its 4D version, termed as 4D LF or simply LF. It quantifies the intensity of static and monochromatic light rays propagating in half space. In this representation, the LF ray positions are indexed either by their Cartesian coordinates on two parallel planes, the so-called two-plane parameterization $L(u,v,s,t)$, or by their one plane and direction coordinates $L(u,v,\phi,\theta)$ ~\cite{liang2011light}.

Consider camera with image plane $(u,v)$  and focal distance $f$ moving along the $(s,t)$ plane. This is an important practical consideration, which associates the parameterizing planes with LF acquisition and multiview imagery and relates LF sampling with discrete camera positions and a discrete camera sensor. The case is illustrated in  \figurename~\ref{fig:epi}~(a) where the $z$ axis represents the scene depth and the plane axes $s$ and $u$ are considered perpendicular to the figure and omitted for simplicity. Constraining the vertical camera motion by fixing $s=s_0$ and moving the camera along the $t$-axis, leads to so-called horizontal parallax only (HPO) multiview acquisition. Images captured by successive camera positions $t_1,t_2,\ldots$ can be stacked together which is equivalent to placing the $t$-axis perpendicular to the $(u,v)$  plane. The corresponding LF $L(u,v,s_0,t)$ is illustrated in  \figurename~\ref{fig:epi}~(b).

\subsection{EPI Representation and Sampling Requirements}\label{sec:epi}

The LF data organization as in  \figurename~\ref{fig:epi}~(b) leads to the concept of EPIs pioneered by Bolles et al. in~\cite{bolles1987epipolar}. Assume an ideal horizontal camera motion (or, equivalently, perfectly rectified perspective images). Gathering image rows for fixed $u=u_0$ along all image positions forms an LF slice $E(v,t)=L(u_0,v,s_0,t)$. Such LF slice is referred to as EPI and is given in \figurename~\ref{fig:epi}~(c). In the EPI, relative motion between the camera and object points manifests as lines with depth depending slopes. Thus, EPIs can be regarded as an implicit representation of the scene geometry. In comparison with regular photo images, an EPI has a very well defined structure. Any visible scene point appears in one of the EPIs as a line whose slope depends on the distance of the point from the capture position and the measured intensity over the line reflects the intensity of emanated light from that scene point. The Lambertian reflectance model (any point in the scene emanates light in different direction with same intensity) leads to an EPI with even more definitive structure -- each line in the EPI has a constant intensity proportional to the intensity of the point. For a scene point at depth $z_0$ measured from the capture plane $(s_0,t)$, the disparity in the image plane $(u_0,v)$ between two cameras positioned at $t_1$ and $t_2$ is~\cite{chai2000plenoptic}
\[
\Delta v=v_2 - v_1= \frac{f}{z_0}(t_2 - t_1)=\frac{f}{z_0}\Delta t,
\]
where $f$ is the camera focal distance. This is illustrated by the red lines in  \figurename~\ref{fig:epi}~(a), which show a point projected on cameras at $t_1$ and $t_2$. The same point appears as the red line in \figurename~\ref{fig:epi}~(c).

By assuming a horizontal sampling rate $\Delta v$ satisfying the Nyquist sampling criterion for scene's highest texture frequency, one can relate the required camera motion step (sampling) with the scene depth. For given $z_{min}$ the sampling rate $\Delta t$ should be such that

\begin{equation}
\label{eqn:deltt}
\Delta t \leq \frac{z_{min}}{f} \Delta v
\end{equation}
in order to ensure maximum 1 pixel disparity between nearby views ~\cite{lin2004geometric},~\cite{chai2000plenoptic}. \figurename~\ref{fig:epi}~(d) shows the frequency domain support of a densely sampled EPI, which is of bow-tie shape. The baseband (in green) is limited by the minimum and maximum depth and its replicas are caused by the sampling rates $\Delta v$ and $\Delta t$. In Fourier domain depths are transformed to lines, i.e. the frequency support of all scene points at a certain depth $z_0$, which in EPI appear as lines with same slope, is confined to a line (the yellow  line in  \figurename~\ref{fig:epi}~(d)). By selecting equality for $\Delta t$ in Eq. (\ref{eqn:deltt}), we effectively place the $z_{min}$ line at 45 degrees in the frequency domain plane. This maximizes the baseband support, which helps in designing reconstruction filters. In particular, simple separable filters (e.g. linear interpolators) can be used. 

\subsection{Motivation}\label{sec:motivation}

Our problem in hand is to reconstruct densely sampled EPIs (and thus the whole LF) from their decimated and aliased versions produced by a higher camera step $\Delta t$. The problem is illustrated in  \figurename~\ref{fig:epi}~(e). The figure shows a case, where a densely sampled EPI has been decimated by a factor of 4, which means that every 4th row has been retained while the others have been zeroed (see also \figurename~\ref{fig:epirows}~(b) for illustration of subsampling in EPI domain). As seen in the figure, aliased replicas (gray) and baseband (green) overlap, hence a band-limited reconstruction is infeasible with a classical filtering method. The work~\cite{chai2000plenoptic} has specified requirements for the LF sampling density for given $z_{min}$ and $z_{max}$ in order to allow a global band-limited reconstruction. Reconstruction of more complex scenes (e.g. piecewise-planar or tilted-plane) would require additional information about scene depth and depth layering~\cite{pearson2013plenoptic},~\cite{chai2000plenoptic}. For real scenes it is natural to assume that objects are distributed at a finite, rather small number of depths. In our approach, we aim at implicitly determining those sparse depth layers by analyzing the given aliased EPIs in frequency domain using depth guided filters. This is equivalent to applying a proper frequency plane tiling. The case in  \figurename~\ref{fig:epi}~(e) is further analyzed in  \figurename~\ref{fig:epi}~(f), which highlights a frequency plane tiling by 4 depth layers, with 1 px disparity range in each layer. If those depth layers are given, they are sufficient to support interpolating the EPIs without aliasing artifacts. Our aim is thus to implicitly obtain those depth layers by transform-domain analysis implying the scene's depth sparse distribution. Furthermore, by an additional dyadic separation of the frequency plane, i.e. a multiresolution analysis, one can process each region differently and utilize a more efficient analysis tool.  \figurename~\ref{fig:epi}~(g) illustrates a wavelet based separation of the frequency plane for the same aliased EPI. It is easy to notice, that the $L_1$ region does not contain any aliasing. Therefore by applying a low-pass filter corresponding to the $L_1$ region on the aliased EPI will reconstruct the desirable densely sampled EPIs frequencies in that region. In other words, the procedure of low-pass filtering followed by decimation can be interpreted as increasing the pixel size, which directly decreases the disparity between the given rows. In this manner, fewer required depth layering directions will have to be distinguished from each other in order to efficiently reconstruct the full EPI. Based on the above discussion, the desirable frequency plane tiling with elemental filters for the case of densely sampled EPI reconstruction from its 4th row subsampled version is given in  \figurename~\ref{fig:epi}~(h). The construction of such set of filters is closely related to the definition and constructive reconstruction of shearlet frames as presented in the next section.

\section{Epipolar-plane image in transform domain}\label{sec:epitd}

\begin{figure*}[t]
\centering
\includegraphics[width=6.5in]{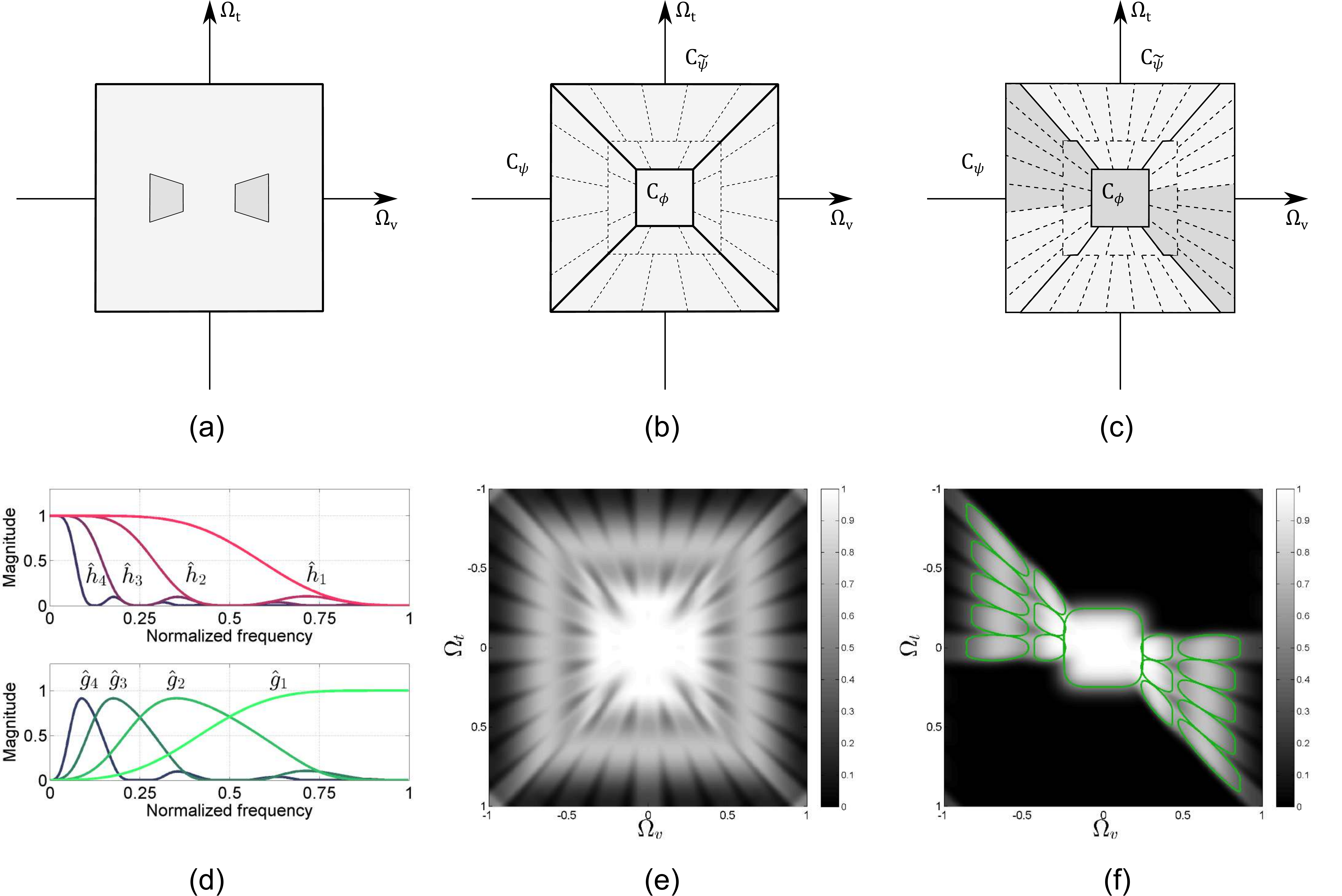}
\caption{(a) Shearlet support in Fourier domain; (b) Frequency plane tilting by shearlet transform. $C_\psi$,$C_{\tilde{\psi}}$  are cone-like regions and $C_\phi$ is low-frequency region; (c) Desirable frequency domain tilting by proposed reconstruction algorithm. Gray color region includes transform elements used for reconstruction; other transform elements are not associated with valid shear values (disparities) in EPI. (d) Frequency responses of the scaling and wavelet filters $h_j,g_j,j=1,…,4$. (e) $\hat{\Psi}^d$ corresponding to constructed shearlet transform for $J=2$. (f) Frequency domain support of shearlet transform elements used in reconstruction algorithm corresponding to gray color region in (c). Green color regions in (f) represent significant parts of transform elements’ support in frequency domain.}
\label{fig:shear}
\end{figure*}

\subsection{Directional sensitive transforms}

Consider a class of piecewise-smooth functions $\varepsilon^2 (\mathbb{R}^2)$ (also referred to as cartoon-like images), as discussed in~\cite{candes1999curvelets},
\cite{candes2004new},\cite{kutyniok2012book}. A function $f \in \varepsilon^2 (\mathbb{R}^2)$ consists of two components and has a form $f=f_0 + f_1 \chi_B$, where $f_0,f_1$ are $C^2$-smooth with support in $[0,1]^2$ and $\chi_B$ is characteristic function of a set $B\subset [0,1]^2$ with bound $\delta B$ being a closed $C^2$-curve with bounded curvature. The problem of reconstructing a function from $\varepsilon^2 (\mathbb{R}^2)$ space using its given incomplete measurements can be addressed through sparse approximation using some appropriately constructed transform. The quality of the representation performance of the $f \in \varepsilon^2 (\mathbb{R}^2)$ in given frame is described by the asymptotic decay speed of the $L^2$ error of the approximation obtained using only $N$ largest coefficients of the frame decomposition. Wavelet-domain decomposition has significant drawback in representing the considered $\varepsilon^2 (\mathbb{R}^2)$  function space. For wavelets, the approximation error rate is $O(N^{-1})$, where $N$ is the number of best elements of a wavelet in a decomposition used for function representation~\cite{kutyniok2012shearlets}. In comparison, adaptive triangle based approximation of the cartoon-like images provides $O(N^{-2})$ approximation rate ~\cite{donoho2001sparse}, where $N$ is the number of triangles used for image representation. This result provides the desirable sparse approximation rate to be achieved using frame based representation for piecewise-smooth functions. In order to provide better approximation than the wavelet transform, the desirable transform should provide a good directional sensitivity due to approximation of singularities distributed over the $C^2$-smooth curve $\delta B$ which is the border between smooth image pieces. Several frames and corresponding transforms have been constructed for sparse representations, among them, tight curvelet frames by Candes and Donoho~\cite{candes2004new} and countourlets by Do and Vetterli~\cite{do2005contourlet}. Going back to the case of study, namely the EPI, one can observe that the anisotropic property of the EPI is caused by a shear transform. This naturally leads to the idea of using a transform constructed with the same property, namely the  shearlet transform.

The optimal sparse approximation property of the tight shearlet frame has been studied in~\cite{guo2007optimally}. Similar results for compactly supported shearlet frame have been reported in~\cite{kutyniok2011compactly}. Both types of shearlet frame construction provide an optimal sparse approximation of $f\in\epsilon^2(\mathbb{R}^2)$, in the sense that $N$-term approximation $f_N$ constructed by keeping $N$ largest coefficients of the frame decomposition satisfies
\[
\norm{f-f_N}^2_2=O(N^{-2}(\log N)^3).
\]

\subsection{Compactly supported shearlets}

The general definitions of a shearlet system and shearlet group can be found in~\cite{kutyniok2012book}, which contains also all necessary conditions about discretization of the parameters to construct the shearlet frame of functions. We will describe in more detail the so-called cone-adapted shearlet system, which is appropriate to construct desirable separation of frequency domain.
Let us consider a scaling function $\phi \in L^2 (\mathbb{R}^2)$ and shearlets $\psi,\tilde{\psi} \in L^2 (\mathbb{R}^2)$. For $c=(c_1,c_2 )\in \mathbb{R}_+^2$ the density of the (regular) cone-adapted discrete shearlet system $SH(\phi,\psi,\tilde{\psi})$ is the set of functions for parameters $j\geq 0, |k|\leq 2^{\lceil j/2 \rceil},m \in \mathbb{Z}^2$
\[
SH(c;\phi,\psi,\tilde{\psi})=\Phi(\phi)\cup \Psi(\psi)\cup \tilde{\Psi}(\tilde{\psi}),
\]
where
\begin{IEEEeqnarray*}{lCr}
\Phi(c_1,\phi)=
{\phi_m=\phi(\cdot - c_1 m),m \in \mathbb{Z}^2 }, \\
\Psi(c,\psi)=\left\{\psi_{j,k,m}=
2^{(j+\lfloor j/2 \rfloor)/2 j} \psi(S_k A_{2^j}\cdot - M_c m)\right\}, \\
\tilde{\Psi}(c, \tilde{\psi})=
\left\{\tilde{\psi}_{j,k,m}=2^{\frac{j+\lfloor j/2 \rfloor }{2}j}\tilde{\psi}(S_k^T \tilde{A}_{2^j}\cdot - \tilde{M}_c m)\right\},
\end{IEEEeqnarray*}
and 
$A_{2^j}=\begin{pmatrix} 2^j & 0 \\ 0 & 2^{\lfloor {j/2} \rfloor} \end{pmatrix}$,
$\tilde{A}_{2^j}=\begin{pmatrix} 2^ {\lfloor {j/2} \rfloor} & 0 \\ 0 & 2^j \end{pmatrix}$
 are parabolic scaling matrices, 
$S_s=\begin{pmatrix} 1 & s\\ 0 & 1\end{pmatrix}$
 is a shearing matrix, and 
$M_c=\begin{pmatrix} c_1 & 0 \\ 0 & c_2 \end{pmatrix}$, 
$\tilde{M}_c=\begin{pmatrix} c_2 & 0 \\ 0 & c_1 \end{pmatrix}$
 are sampling densities of translation grid. The transform maps $f\in L^2 (\mathbb{R}^2)$ to the sequence of coefficients
\[
f \rightarrow \langle f,\tau \rangle, f \in SH(c;\phi,\psi,\tilde{\psi}).
\]

The function $\Phi(\phi)$ handles the region close to the origin $C_\phi$ and $\Psi(\psi),\tilde{\Psi}(\tilde{\psi})$ handle the $C_\psi,C_{\tilde{\psi}}$  cones as illustrated in \figurename~\ref{fig:shear}~(b). This is accomplished by appropriately selecting the generator functions $\phi,\psi,\tilde{\psi}$. More details about the construction and sufficient conditions for forming frame of functions from the cone-adapted shearlet system can be found in~\cite{kutyniok2012book}. In addition, it is also desirable to have compact support of the shearlet frame elements that is important for the algorithm considered in this paper. In order to construct a compactly supported shearlet system, different approaches have been considered \cite{kutyniok2011compactly}, \cite{lim2013nonseparable}. In ~\cite{kutyniok2011compactly}, a compactly supported shearlet has been constructed in spatial domain by selecting 2D separable functions as generators
\begin{IEEEeqnarray*}{lCr}
\label{eq:sep}
\psi(x_1,x_2)=\psi_1 (x_1) \phi_1 (x_2), \\
\phi(x_1,x_2)=\phi_1 (x_1) \phi_1 (x_2), \IEEEyesnumber \\ 
\tilde{\psi}(x_1,x_2)=\psi(x_2,x_1),
\end{IEEEeqnarray*}
where $\phi_1(x_1)$, $\psi_1(x_1)$ are 1D scaling and wavelet functions. Henceforth, we will denote $\hat{f}(\xi)$ as the Fourier transform of $f(x)$. Under certain conditions on $\psi_1, \phi_1$ the corresponding expansion $SH(c;\phi, \psi, \tilde{\psi})$ form a frame and the constructed elements of the frame are compactly supported in spatial domain. Choosing separable shearlet generator $\psi (x_1,x_2)$ is not efficient. Significant overlap between $\supp(\hat{\psi}_{j,k,m})$ and $\supp(\hat{\psi}_{j,k+1,m})$ makes shearlet frame over redundant and with bad directional selectivity \cite{lim2013nonseparable}.

A construction of compactly supported shearlets by selecting non-separable $\psi$ $2D$ directional filter has been presented in~\cite{kutyniok2011compactly}. Using such shearlet generator provides better covering of frequency plane by transform elements and has better directional selectivity. In order to present the construction and implementation of the proposed shearlet transform we follow the implementation of the digital non-separable 2D shearlet transform, associated with compactly supported shearlets as presented in~\cite{kutyniok2014shearlab}.

Consider a multiresolution analysis with wavelet and scaling function $\psi_1,\phi_1\in L^2 (R)$ given as
\begin{IEEEeqnarray*}{lCr}
\phi_1 (x_1)=\sum_{n_1\in \mathbb{Z}} h(n_1 ) \sqrt{2} \phi_1 (2 x_1 - n_1) \\
\psi_1 (x_1)=\sum_{n_1\in \mathbb{Z}} g(n_1 ) \sqrt{2} \phi_1 (2 x_1 - n_1).
\end{IEEEeqnarray*}
The nonseparable shearlet generator is defined as
 \[
\hat{\psi}(\xi_1,\xi_2 )=P(\xi_1/2,\xi_2 ) \hat{\psi}_1 (\xi_1 ) \hat{\phi}_1 (\xi_2),
 \]
where the trigonometric polynomial $P$ is a 2D directional fan filter~\cite{do2005contourlet} which is used to approximate the 2D non-separable filter with essential support in frequency domain bounded within the region shown in~\figurename~\ref{fig:shear}~(a).
 
In order to present discrete shearlet transform of the continuous function $f(x),x\in \mathbb{R}^2$ we assume  that for some sufficiently large $J\in \mathbb{N}$, the function can be represented using its discrete samples $f_J (n),n \in \mathbb{Z}^2$
\[
f(x)=\sum_{n \in \mathbb{Z}^2}{f_J(n) 2^J \phi(2^J x-n)},
\]
where $\phi(x)$ is defined as in Eq.~(\ref{eq:sep}). The particular choice of $J$ depends on the given input data and will be discussed in Section~\ref{sec:recalg}.

\subsection{Modified shearlet transform for EPI}

The choice of $A_{2^j}$ as a parabolic scaling matrix, has been motivated by the aim to construct best approximation of function with singularities over parabolic curves. For the EPI one can observe singularities distributed over straight lines. Therefore, a better choice for the scaling transform is given by
\[
A_{2^j}=\begin{pmatrix} 2^j & 0 \\ 0 & 2^{-1} \end{pmatrix}.
\]
This choice supports the desirable number of shears in each scale and provides scaling only by one axis. It can be considered as a special case of a more general shearlet transform called universal shearlet \cite{kutyniok2012book}, \cite{kutyniok2014shearlab}. Furthermore the shearlet system for $\Psi(\psi)$ consists of the functions
\[
\Psi(c,\psi)={\psi_{j,k,m},|k|\leq 2^{j+1},j=0,\ldots,J-1},
\]
where
\begin{equation}
\label{eq:psijkm1}
\psi_{j,k,m} (x)=2^{j/2} \psi \left(S_k A_{2^j} x - M_{c_j} m \right),
\end{equation}
and $c_j=(c_1^j,c_2^j)$ are sampling constants for translation. Easy to notice
\begin{equation}
\label{eq:psijkm2}
\psi_{j,k,m} (x) = \psi_{j,0,m} \left(S_{\frac{k}{2^{j+1}}} x \right).
\end{equation}

Following the same methodology as in~\cite{kutyniok2014shearlab}, it can be shown that the digital filter corresponding to $\psi_{j,0,m}$ has the form
\begin{equation}
\label{eq:psij0d}
\psi_{j,0}^d (m)=\left( p_j \ast \left( g_{J-j} \otimes h_{J+1} \right) \right) (m),
\end{equation}
where $\otimes$ denote tensor  product such that $\left( g_{J-j} \otimes h_{J+1} \right)(m) = g_{J-j}(m_1)h_{J+1}(m_2)$, $\left\{p_j (n)\right\}_{n\in \mathbb{Z}}$ are the Fourier coefficients of the trigonometric polynomial 
$P(2^{J-j-1} \xi_1,2^{J+1} \xi_2)$,
$\{h_j (n)\}_{n\in \mathbb{Z}}$ and $\{g_j (n)\}_{n\in \mathbb{Z}}$ are the Fourier coefficients of the respective trigonometric polynomials
\begin{IEEEeqnarray*}{lCr}
\hat{h}_j (\xi)=\prod_{k=0,\ldots,j-1} {\hat{h}(2^k \xi)}, \\
\hat{g}_j (\xi)=\hat{g}(2^{j-1} \xi) \hat{h}_{j-1} (\xi)	
\end{IEEEeqnarray*}
and $\hat{h}_0 \equiv 1$. \figurename~\ref{fig:shear}~(d) illustrates the frequency responses of the digital filters $h_j, g_j$ for $j=1,\ldots,4$.

The shear transform $S_{k2^{-j}},j\in \mathbb{N},k\in \mathbb{Z}$ does not preserve the regular grid $\mathbb{Z}^2$, therefore its digitalization is not trivial. The solution to the problem presented in~\cite{lim2013nonseparable}, is to refine the $\mathbb{Z}^2$ grid along $x$-axis by a factor $2^j.$ In that case, the grid $2^{-j}\mathbb{Z} \times \mathbb{Z}$  is invariant under the $S_{k2^{-j}}$ transform. Thus, for an arbitrary $r\in l^2 (\mathbb{Z}^2)$, the shear transform $S_{k2^{-j}}$ can be implemented as a digital filter
\begin{equation}
\label{eq:pfull}
S_{k2^{-j}}^d (r)=\left((2^j r_{\uparrow 2^j} \ast_1 \tau_j )
(S_k \cdot) \ast_1 \bar{\tau}_j\right)_{\downarrow 2^j}
\end{equation}
where $\tau_j$ represents a digital low-pass filter with normalized cutoff frequency at $2^{-j}$.

Based on previous results from Eq.~(\ref{eq:psijkm1}),~(\ref{eq:psijkm2}),~(\ref{eq:psij0d}),~(\ref{eq:pfull}) and proper choice of $c_j$ it can be shown that digital filter corresponding to $\psi_{j,k,m}$ has a form 
\[
\psi_{j,k}^d=(S_{k2^{-(j+1)}}^d (p_j\ast g_{J-j}\otimes h_{J+1} ))(m).
\]
A digital filter $\psi^d$ corresponding to separable elements of the transform $\phi_m$, has a form $\phi^d=(h_J\otimes h_J )(m)$.

Based on the above, a direct shearlet transform associated with set of elements $\Psi(c;\psi)$ and corresponding to frequency plane region $C_\psi$ is defined as follows
\[
{DST}_{j,k,m} (f_J )=(f_J\ast \overline{\psi_{j,k}^d} )(m),
\]
where $j=0,\ldots,J-1,|k| \leq 2^{j+1},m\in \mathbb{Z}^2$.

In order to calculate the inverse transform we need to construct the dual frame. This can be done based on the shift invariance property of the shearlet frame. First we set
\[
\hat{\Psi}^d = |\hat{\phi}^d|^2 + \sum_{j=0,\ldots,J-1} 
\sum_{|k| \leq 2^{j+1}}(|\hat{\psi}_{j,k}^d |^2+|\hat{\tilde{\psi}}_{j,k}^d |^2) .
\]
Then, the dual shearlet filters are defined as follows
\[
\hat{\phi}^d=\frac{\hat{\phi}^d}{\hat{\Psi}^d},
\hat{\gamma}_{j,k}^d=\frac{\hat{\psi}_{j,k}^d}{\hat{\Psi}^d},
\hat{\tilde{\gamma}}_{j,k}^d=\frac{\hat{\tilde{\psi}}_{j,k}^d}{\hat{\Psi}^d}.
\]

The constructed frame guarantees stable reconstruction using the dual frame, if $A\leq \hat{\Psi}^d \leq B$ conditions are satisfied for some finite bounds $0<A,B<\infty$ ~\cite{mallat2008wavelet}. An illustration of the obtained $\hat{\Psi}^d$ for $J = 2$, is presented in \figurename~\ref{fig:shear}~(e). In this case, the upper and lower bounds are numerically found to be $0.03<\hat{\Psi}^d<1.03$.  The reconstruction formula is given by 
\begin{IEEEeqnarray*}{l}
f_J=(f_J\ast \bar{\phi}^d) \ast \phi^d + 
\sum_{j,k} {(f_J \ast \bar{\psi}_{j,k}^d) \ast \gamma_{j,k}^d}+
\qquad \\
\IEEEeqnarraymulticol{1}{r}{\sum_{j,k} {(f_J \ast \bar{\tilde{\psi}}_{j,k}^d) \ast \tilde{\gamma}_{j,k}^d}.}
\end{IEEEeqnarray*}

We are interested only in transform elements where the shearing has positive sign, i.e. $0\leq k\leq 2^{j+1}$, such that the corresponding transform elements are covering the frequency domain region highlighted by gray in \figurename~\ref{fig:shear}~(c).  Therefore, we use the direct transform $S$ for discrete values $f_J$ and  $ j=0,\ldots,J-1, k=0,\ldots,2^{j+1}, m\in\mathbb{Z}^2$ defined as
\[
S(f_J)=\left\{
c_{j,k}(m)=(f_J \ast \bar{\psi}_{j,k}^d)(m),
c_0(m)=(f_J \ast \bar{\phi}^d)(m)
 \right\}.
\]
Respectively, the inverse transform $S^*$ is defined as
\[
S^* \left( \left\{ c_{j,k}, c_0 \right\} \right)=
\sum_{\substack{j=0,\ldots,J-1\\ k=0,\ldots,2^{j+1}}}
(c_{j,k}\ast \gamma_{j,k}^d)(m) +(c_0\ast \phi^d )(m).
\]

The frequency-domain support of the elements selected from the frame in~\figurename~\ref{fig:shear}~(e) is shown in~\figurename~\ref{fig:shear}~(f). 

\section{Reconstruction algorithm}\label{sec:recalg}

\begin{figure}[t]
\centering
\includegraphics[width=3.5in]{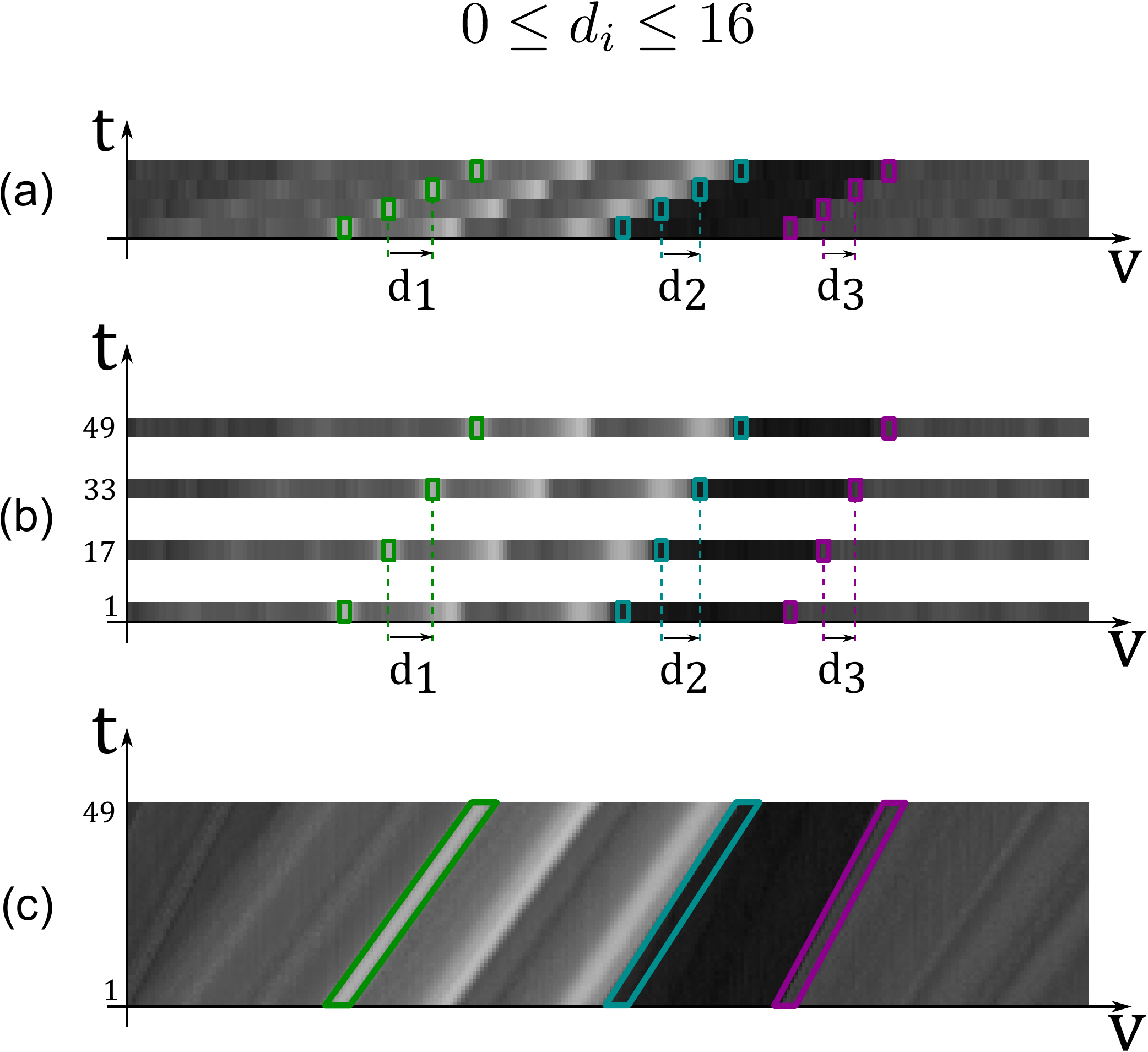}
\caption{The given 4 views with maximal disparity 16 px between consecutive views are interpreted as every 16th view in the target densely sampled light field. (a) EPI for coarsely sampled light field over $t$-axis; (b) corresponding partially defined densely sampled EPI;  (c) ground truth densely sampled EPI. Three different points from given input images forming traces are highlighted in the coarsely (a) and densely (c) sampled EPIs. Only in (c) they are revealed as a straight lines.}
\label{fig:epirows}
\end{figure}

In this section we present an LF reconstruction algorithm utilizing an EPI sparse representation in shearlet domain. Usually, a setup of uniformly distributed, parallel positioned and rectified cameras is used for capturing a 3D scene. The horizontal parallax between views limits the motion associated with the depth of the objects in horizontal axis only. This allows us to perform intermediate view generation over EPI independently. In order to formulate the reconstruction algorithm in discrete domain we assume that the starting coarse set of views is downsampled version of the unknown densely sampled LF we try to reconstruct. The uniformly distributed cameras imply the possibility of estimating a common upper bound $d_{max}$ for disparities between nearby views. Thus, the given coarse set of views are regarded as taken at each $d_{max}=\lceil d_{max} \rceil$-th view of a densely sampled LF. 

Thus, in every densely sampled EPI, all unknown rows should be reconstructed assuming given every $d_{max}$-th row. An example is presented in~\figurename\ref{fig:epirows}~(a), where EPI representation of four views with 16 px disparity is given. Therefore, the targeted densely sampled EPI is to be constructed in such a way that the available data will appear in rows with 16 px distance (\figurename\ref{fig:epirows}~(b)). \figurename\ref{fig:epirows}~(c) shows the same rows with respect to the fully reconstructed EPI, where successive rows appear at disparity less or equal to 1 px. EPI lines are not distinguishable in ~\figurename\ref{fig:epirows}~(a). The lines start to form when the views are properly arranged, as in~\figurename\ref{fig:epirows}~(b), and they get fully reconstructed in the densely sampled EPI.

\begin{figure}[!t]
\centering
\includegraphics[width=3.5in]{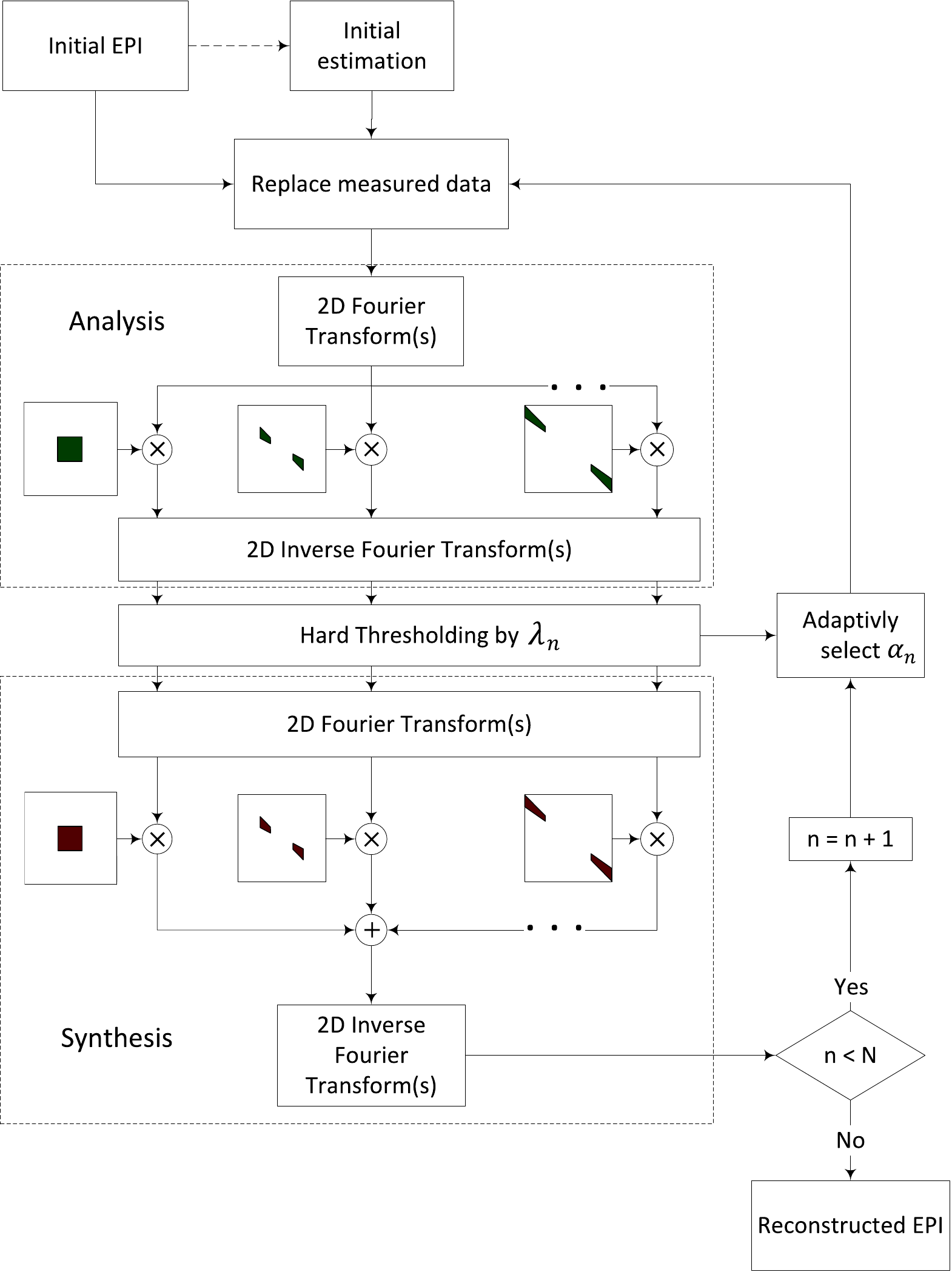}
\caption{Diagram of the EPI reconstruction algorithm.}
\label{fig:diag}
\end{figure}

Without loss of generality we assume that the densely sampled EPI is a square image denoted by $y^*\in \mathbb{R}^{N\times N}$ where $N=m d_{max}$ and $m$ is a number of available views. Given the samples $y\in \mathbb{R}^{N\times N}$ of the $y^*$ obtained by
\begin{equation}
\label{eq:yHy}
y(i,j)=H(i,j)y^*(i,j),
\end{equation}
where $H\in \mathbb{R}^{\mathbb{N}\times \mathbb{N}}$ is a measuring matrix, such that $H(k d_{max},\cdot) = 1, k = 1,\ldots,m$ and 0 elsewhere. The measurements  $y$ form an incomplete EPI where only rows from the available images are presented, while everywhere else EPI values are $0$. Eq.~(\ref{eq:yHy}) can be rewritten in the form $y=Hy^*$ by lexicographically reordering the variables $y, y^*\in \mathbb{R}^{\mathbb{N}^2},H \in \mathbb{R}^{\mathbb{N}^2\times \mathbb{N}^2}$. The shearlet analysis and synthesis transforms are defined as 
$
S:\mathbb{R}^{\mathbb{N} \times \mathbb{N}} \rightarrow \mathbb{R}^{\mathbb{N} \times \mathbb{N}\times \eta}, 
S^*:\mathbb{R}^{\mathbb{N} \times \mathbb{N}\times \eta} \rightarrow \mathbb{R}^{\mathbb{N} \times \mathbb{N}},
$
where $\eta$ is the number of all translation invariant transform elements.

The reconstruction of $y^*$ given the sampling matrix $H$ and the measurements $y$ can be cast as an inpainting problem, with constraint to have solution which is sparse in the shearlet transform domain, i.e.
\begin{equation}
\label{eq:prob}
x^*=\argmin_{x \in \mathbb{R}^{N\times N}}{\norm{S(x)}_1}, \text{subject to } y=Hx.
\end{equation}
We make use of the iterative procedure within the morphological component analysis approach, which has been originally proposed for decomposing images into piecewise-smooth and texture parts~\cite{starck2005morphological}, \cite{fadili2009mcalab}. In particular, we aim at reconstructing the EPI $y^*$ by performing regularization in shearlet transform domain. Solution is sought in the form of the following iterative thresholding algorithm
\[
x_{n+1} = S^* \left(T_{\lambda_n}(S(x_n + \alpha_n (y - H x_n)))\right),
\]
where 
$(T_\lambda x)(k)= \left\{
\begin{array}{ll}
x(k), |x(k)| \geq \lambda\\
0,|x(n)|<\lambda
\end{array}
\right.$
 is a hard thresholding operator applied on transform domain coefficients and $\alpha_n$ is an acceleration parameter. The thresholding level $\lambda_n$ decreases with the iteration number linearly in the range $[\lambda_{max},\lambda_{min}]$. After sufficient number of iterations, $x_n \rightarrow x^*$ reaches a satisfying solution of the problem~(\ref{eq:prob}). The diagram of the reconstruction method is given in~\figurename~\ref{fig:diag}.

\begin{figure}[t]
\centering
\includegraphics[width=3.5in]{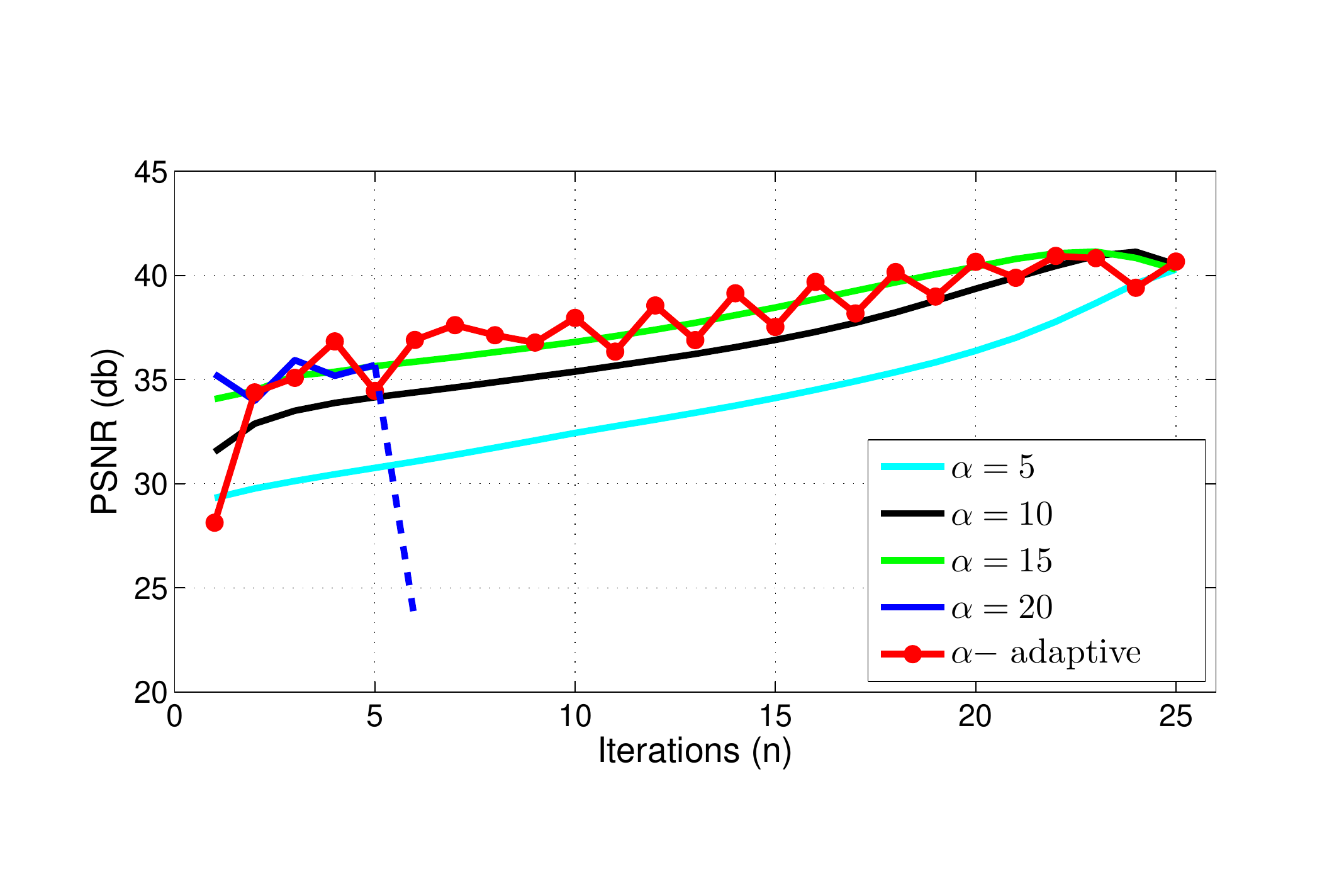}
\caption{Example of reconstruction performance dependence on choice of acceleration coefficients $\alpha_n$. For constant value for all iterations  $\alpha_n=\alpha$, increasing $\alpha$ brings accelerating convergence. After some limit, reconstruction starts to diverge $(\alpha=20)$.}
\label{fig:alpha}
\end{figure}

The rate of convergence is controlled by the parameter $\alpha_n$. For the case of $\alpha_n=1$ the convergence is slow and can be accelerated by selecting $\alpha_n>1$. However, selecting alpha too high can cause instability.  The case is illustrated in~\figurename~\ref{fig:alpha} where convergence speed benefits from increasing yet fixed values $\alpha_n=\alpha$ up to some level where the algorithm starts to diverge. This motivates us to apply an iteration-adaptive selection of the parameter $\alpha_n$. We devise the adaptation procedure in the way as proposed in~\cite{blumensath2010normalized}. Let us define $\Gamma_n$ as the support of $S(x_n)$. The adaptive selection of the acceleration parameter is
\[
\alpha_n = \frac{\norm{\beta_n}_2^2}{\norm{H S^* (\beta_n)}_2^2}
\]
where $\beta_n=S_{\Gamma_n}(y - Hx_n)$ and $S_{\Gamma_n}$ is the shearlet transform decomposition only for coefficients from $\Gamma_n$. The convergence rate for the adaptive selection of the acceleration parameter is illustrated in~\figurename~\ref{fig:alpha}. As can be seen in the figure, the adaptation provides high convergence speed and stable reconstruction.

The initial estimate $f_0$ can be chosen either $0$ everywhere or as the result of a low-pass filtering of the input $y$ using the central separable filter $\phi^d$ only.

As discussed previously we are not obliged to use all general shearlet transform atoms. We favor the use  of atoms which are associated with valid directions in EPI, i.e. only those having support in frequency domain enclosed in the region highlighted in  \figurename~\ref{fig:epi}~(d). An example of such subset is presented in  \figurename~\ref{fig:epi}~(h). The scales of the shearlet transform are constructed in dyadic manner, thus we are choosing $J=\lceil \log_2 d_{max} \rceil $ number of scales. In every scale we choose $2^{j+1}+1$ shears $(j=0,\ldots,J-1)$ to cover the region presented in  \figurename~\ref{fig:epi}~(g) associated with $s_k=\frac{k}{2^{j+1}} ,k=0,\ldots,2^{j+1}$ shears (i.e. disparities).

\section{Evaluation}\label{sec:eval}

\begin{figure}[t]
\centering
\includegraphics[width=3.5in]{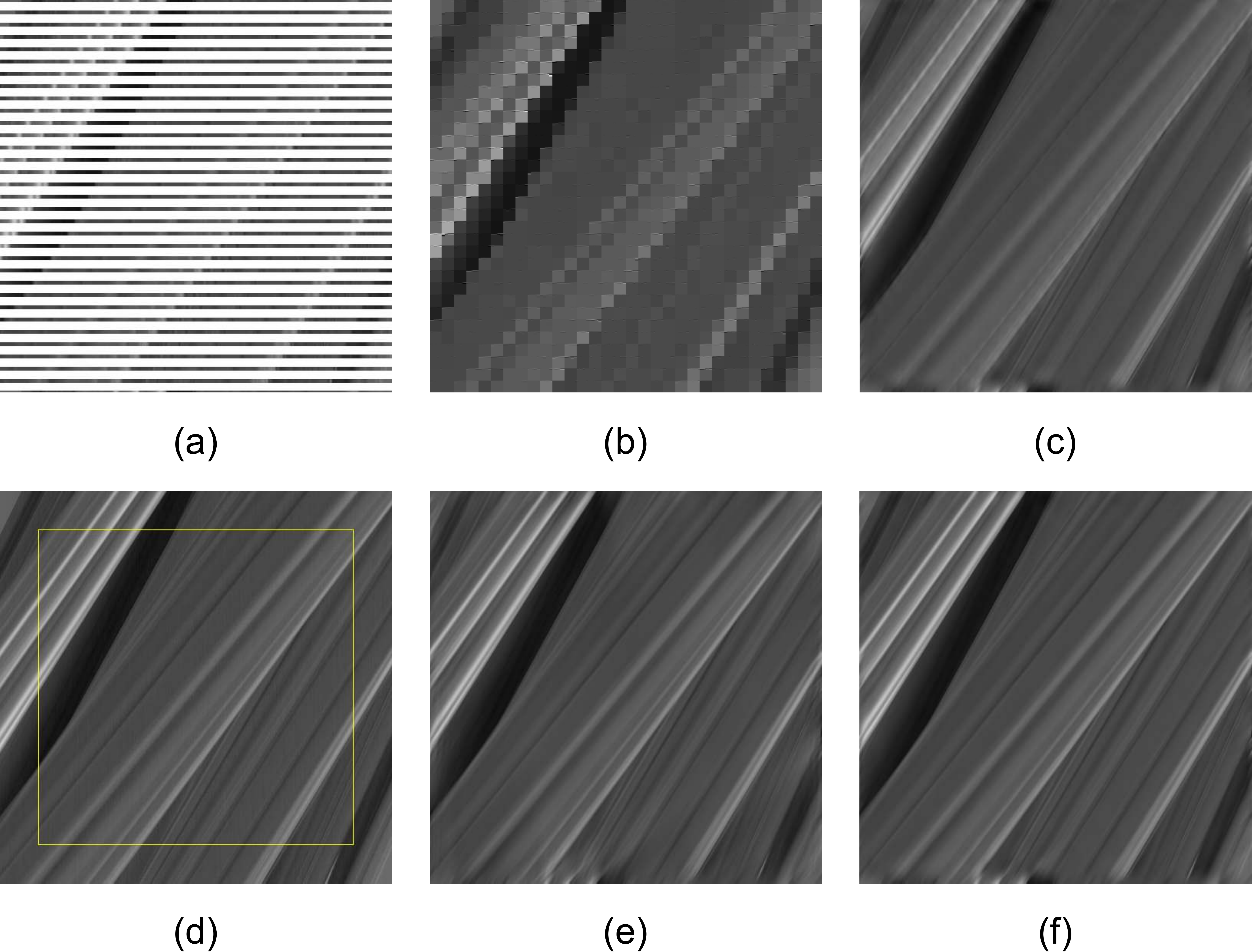}
\caption{(a) Input for reconstructing densely sampled EPI where only every 4th row is available. (d) Densely sampled ground truth EPI. Reconstruction results using different transform are shown as follows (b) Haar 24~dB, (c) shearlab~\cite{kutyniok2014shearlab} 33~dB, (e) FFST~\cite{hauser2012fast} 39.65~dB, (f) proposed 41.57~dB.}
\label{fig:diccomp}
\end{figure}

In this section we evaluate the performance of the proposed method. First, we demonstrate the performance of the reconstruction algorithm with respect to different transforms presented in~\cite{kutyniok2014shearlab},~\cite{hauser2012fast}. Ground truth densely sampled EPI (\figurename~\ref{fig:diccomp}(d)) is obtained using properly generated views of a synthetic scene. Every 16th row has been used as input data for the reconstruction method as in~\figurename~\ref{fig:diccomp}~(a), and interpreted in similar fashion as presented in~\figurename~\ref{fig:epirows}. The obtained reconstruction results are presented in~\figurename~\ref{fig:diccomp}~(b), (c), (e), (f). The reconstruction using Haar wavelet transform is not properly revealing straight lines and the performance is poor. Directional sensitive transforms are showing better reconstruction performance, while the proposed shearlet transform outperform the others. The proposed transform combines two properties, compact support in horizontal direction in spatial domain and tight distribution of transform elements near low-frequency region in frequency plane which affect the reconstruction performance.

Next, we characterize the reconstruction performance for different test sets using leave $N$ out tests. The experimental setup considers downsampled versions of a number of given test multiview sets, where every $N$-th view is kept and the others are dropped. The downsampled versions are used as input to the algorithm, which is supposed to reconstruct all views for the given sets, which have been dropped during the downsampling step. The reconstruction quality is assessed by calculating the PSNR between the original and the reconstructed views. DERS+VSRS~\cite{tanimoto2009depth},~\cite{tanimoto2009view}, is used as a reference algorithm to compare with. DERS (depth estimation reference software) is applied for every three consecutive images in order to estimate disparity map corresponding to the middle view. Using a stack of given images with corresponding estimated disparity maps, the desired intermediate views are generated using VSRS (view synthesis reference software).

\begin{table}[t]
\renewcommand{\arraystretch}{1.3}
\caption{Multiview Data Sets Details}
\label{table:mvd}
\centering
\begin{tabular}{l | M{1.5cm} | M{1.2cm} | M{0.8cm} | M{1.2cm} }
\hline
\bfseries Dataset & \bfseries Resolution & \bfseries Number of views & \bfseries Leave $N$ out & \bfseries $\lceil d_{max} \rceil$\\
\hline\hline
Couch~\cite{kim2013scene} & $2768\times 4020$ & 37 & 2 & 14(RGB) \\
BBB~\cite{kovacs2015bbb} & $320\times 192$ & 91 & 2 & 8(Y),4(UV) \\
Pantomime1~\cite{ToyohiroNagoya} & $640\times 480$ & 81 & 8 & 3(Y),2(UV) \\
Pantomime2~\cite{ToyohiroNagoya} & $640\times 480$ & 81 & 4 & 8(Y),4(UV) \\
Teddy~\cite{scharstein2003high} & $450\times 375$ & 9 & 2 & 17(RGB) \\
Cones~\cite{scharstein2003high} & $450\times 375$ & 9 & 2 & 17(RGB) \\
Truck~\cite{vaish2008new} & $383\times 512$ & $17\times 17$ & 4(2) & 3(RGB) \\
Bunny~\cite{vaish2008new} & $512\times 512$ & $17\times 17$ & 4(2) & 3(RGB) \\
\hline
\end{tabular}
\end{table}

\begin{figure*}[t]
\centering
\includegraphics[width=7in]{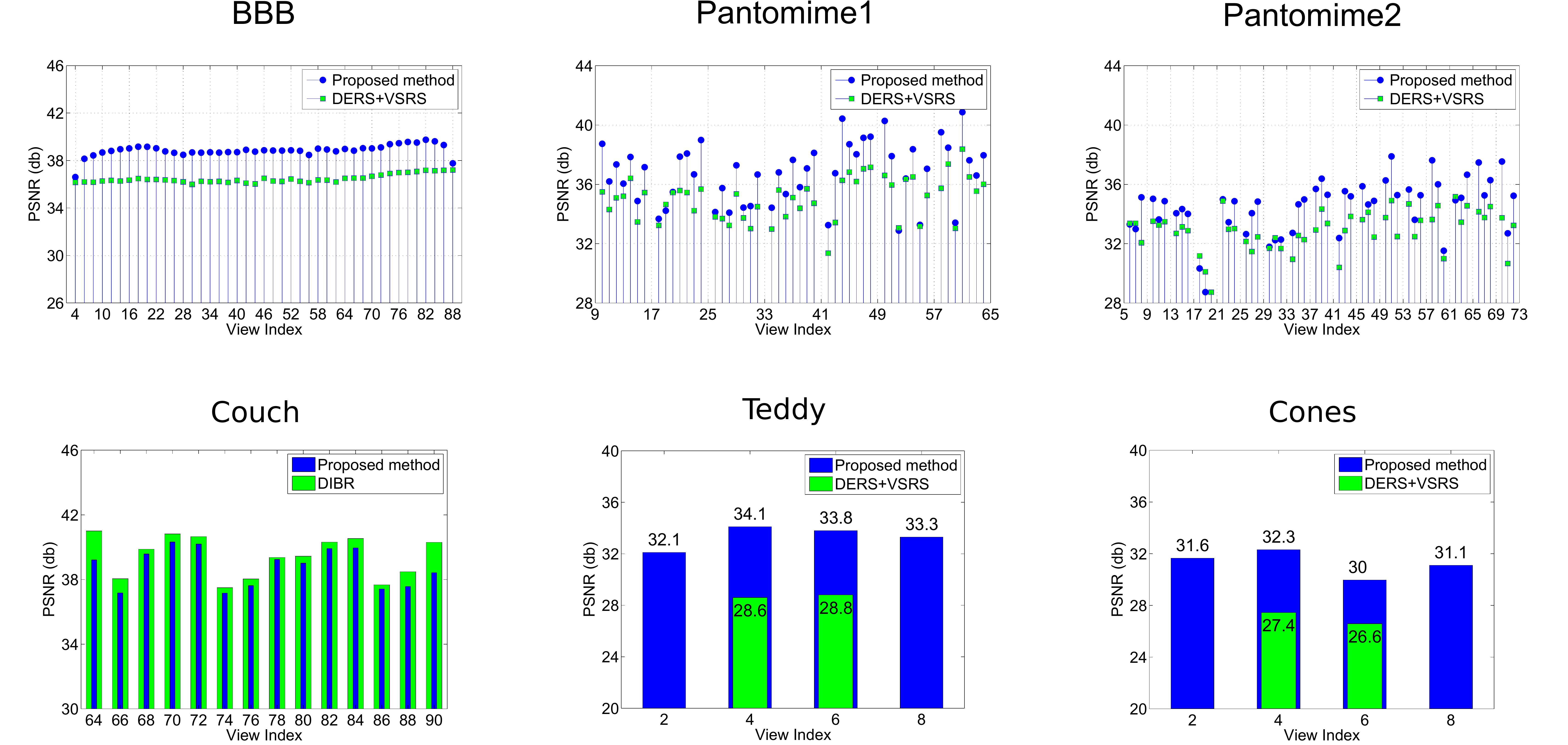}
\caption{Reconstruction results for different multiview datasets, error shown in PSNR for reconstructed views which were not used as input images for the reconstruction algorithm.}
\label{fig:comp}
\end{figure*}

We have used a number of publicly available datasets, as presented in Table~\ref{table:mvd}. The table summarizes also some specifications of the sequences such as spatial resolution and number of views.  In all test cases, our algorithm is applied over all EPIs for reconstructing the missing intermediate views. The adaptive selection of the acceleration parameter, as described in Section~\ref{sec:recalg}, has been applied. Typically, 100 iterations were used per dataset in order to obtain the presented results.

\begin{figure*}[t]
\centering
\includegraphics[width=7in]{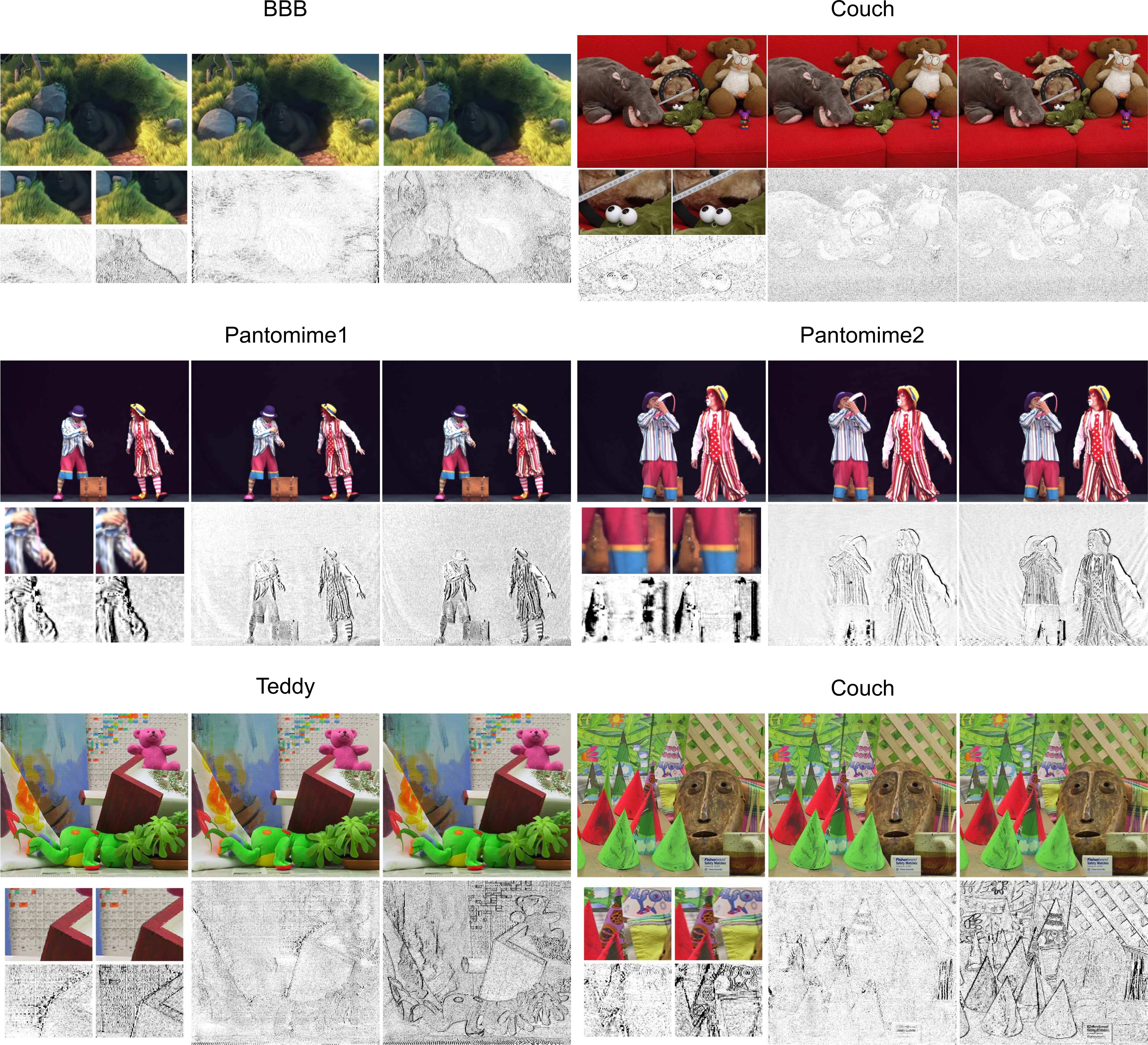}
\caption{Examples of reconstructed views for different datasets. For each dataset the top row shows the ground truth and the reconstruction using the proposed algorithm and a competitor algorithm, consecutively. The bottom row shows zoomed in regions from different reconstructed images and corresponding scaled difference maps.}
\label{fig:compimg}
\end{figure*}

\figurename~\ref{fig:comp} presents the comparative results of reconstruction based on our and the DERS+VSRS algorithms. For efficient implementation of the algorithm, circular convolution was implemented through Fourier transform as presented in the diagram in~\figurename~\ref{fig:diag}. While this makes the procedure fast, in some cases it introduces border effects (e.g. in~\figurename~\ref{fig:compimg} see specifically the Bunny difference map). Our method performs better in all cases but one, notably for the Coach dataset. However, for the Couch dataset instead of DERS, we have used all disparity maps, as already obtained by the algorithm presented in~\cite{kim2013scene}. In the referred algorithm, the disparity maps are estimated using the full set of images, not only the downsampled one. Thus, the depth maps are of higher quality than the one that can be achieved if only the downsampled views are given. The comparison in this case is made in order to quantify the performance of our algorithm against an ideal case of DIBR. A direct comparison of reconstructed views obtained by ours method and the one from \cite{kim2013scene} shows that even in the case of 'unfair' comparison our algorithm reconstructs views with competitive quality. Another observation is that the datasets Pantomime show high variations in reconstruction quality for different views. This property can be observed for both reconstruction algorithms. The cause is in the lack of perfect rectification between views in that dataset. 

We compare our method also against the method presented in~\cite{pearson2013plenoptic} which is another IBR method utilizing depth layering. Both methods show equal performance measured for the Teddy dataset. The PSNRs are averaged over four reconstructed views. The result reported in~\cite{pearson2013plenoptic} shows 33.25~dB, while our method gives 33.33~dB.

\begin{figure}[t]
\centering
\includegraphics[width=3.5in]{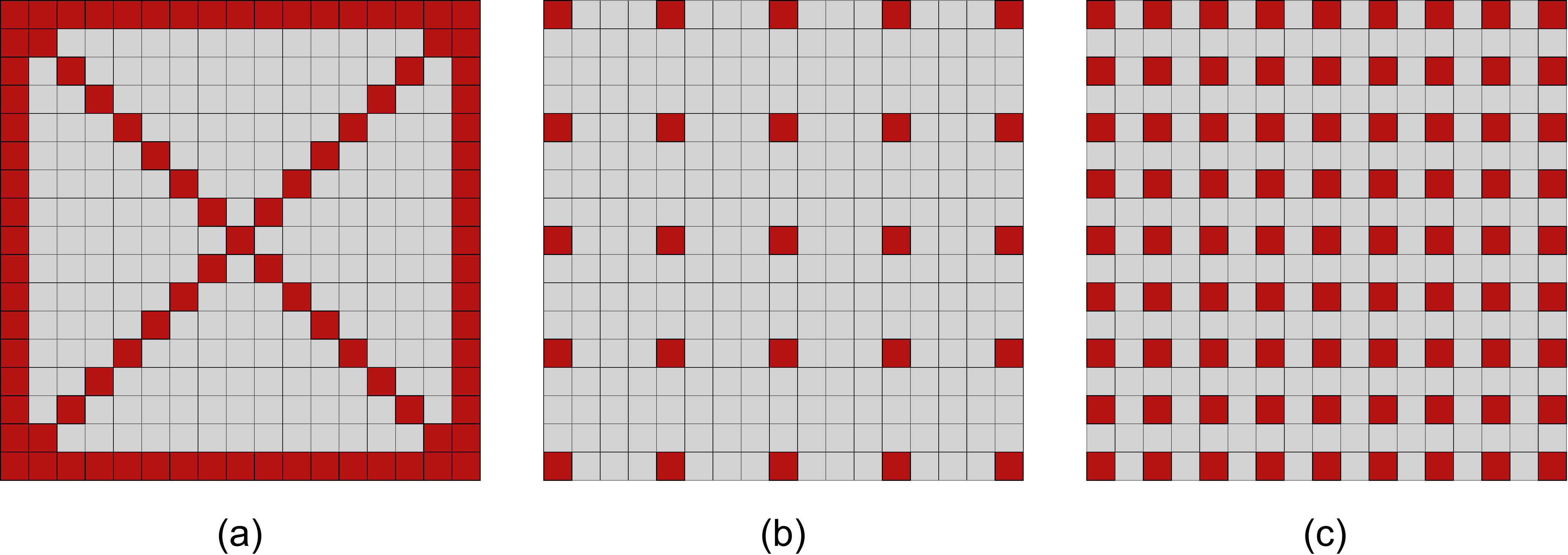}
\caption{Sampling pattern where every rectangle represents one view from the light field consisting of $17\times 17$ views. (a) box and two diagonals pattern consisting of 93 views used for method~\cite{shi2014light}. (b), (c) uniformly decimated setup consisting of $5\times 5$ and $9\times 9$ views respectively. }
\label{fig:pattern}
\end{figure}

\begin{table}[t]
\caption{Light field reconstruction evaluation}
\label{table:lfpsnr}
\centering
\begin{tabular}{l | c | c | c }
\hline
\bfseries Dataset & \bfseries SFFT~\cite{shi2014light} & \bfseries Proposed $5\times 5$ & \bfseries Proposed $9\times 9$\\
\hline\hline
Truck~\cite{vaish2008new} & 36.4~dB & 40.6~dB & 41.2~dB \\
Bunny~\cite{vaish2008new} & 38.8~dB & 37.8~dB & 38.9~dB \\
\hline
\end{tabular}
\end{table}

\begin{figure*}[!th]
\centering
\includegraphics[width=7in]{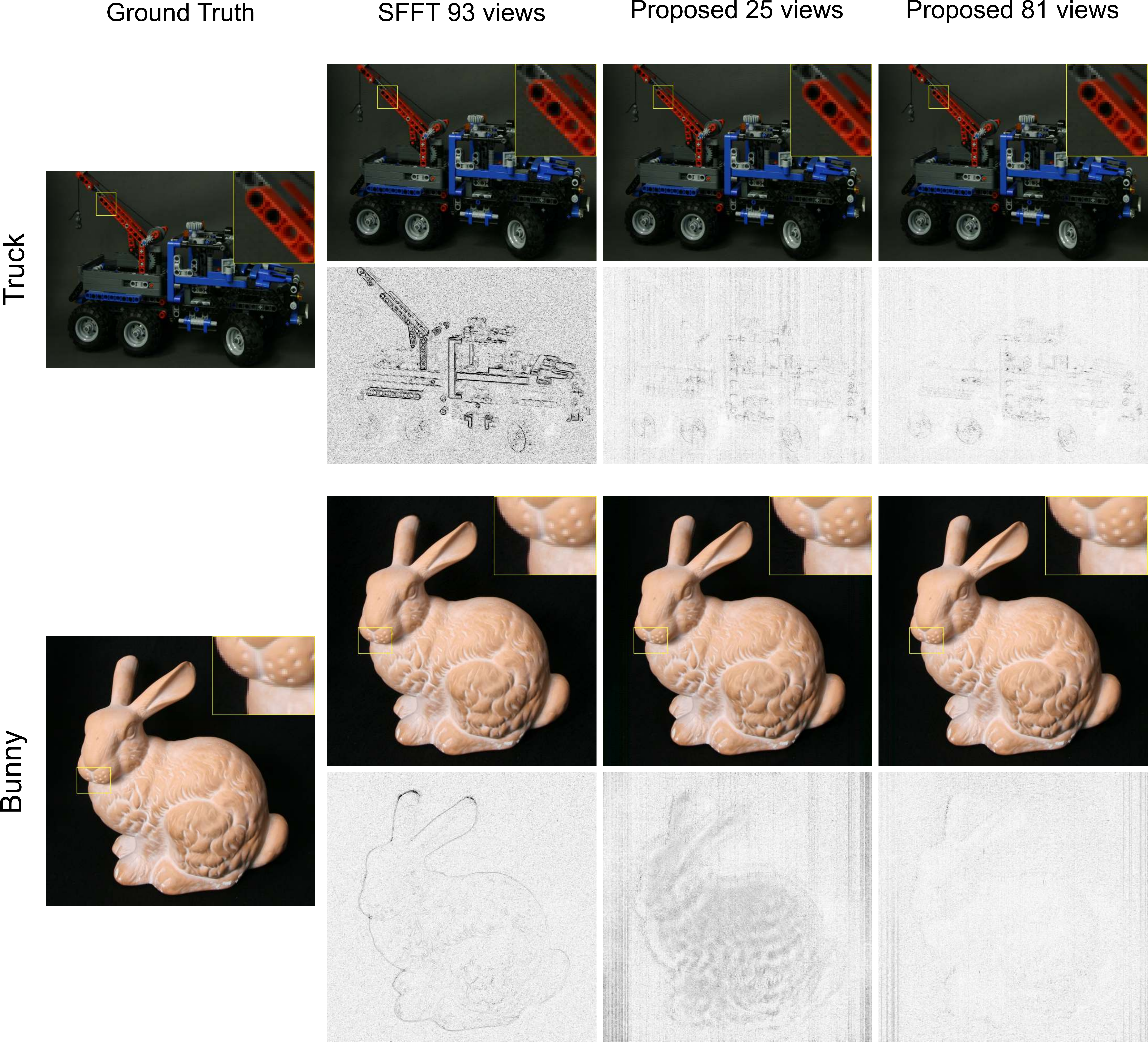}
\caption{Evaluation of the light field reconstruction algorithms using Truck and Bunny datasets. SFFT refers to method presented in~\cite{shi2014light} where required input dataset consist of boarder views and two diagonal sets as shown in~\figurename~\ref{fig:pattern}~(a). Presented two subsequent reconstruction are based on proposed method using input dataset constructed as illustrated in~\figurename~\ref{fig:pattern}~(b),(c).}
\label{fig:complf}
\end{figure*}

Our next tests deal with full parallax imagery. In~\cite{shi2014light}, a method is proposed for LF reconstruction that utilizes sparsity of full parallax LF in continuous Fourier domain. The method can be used for reconstruction for non-Lambertian scenes and it requires a set of views obtained from a set of 1D viewpoint trajectories \cite{shi2014light}. We compared reconstruction results for dataset Bunny and Truck~\cite{vaish2008new} consisting of $17 \times 17$ views, which are representing Lambertian scenes, thus suitable for the proposed and the method in \cite{shi2014light}. In the proposed method the reconstruction is applied for every horizontal and then for every vertical EPI consecutively. Two experiments, one with 25 views and one with 81 views out of 289 has been applied. The method in~\cite{shi2014light} uses 93 views as input. The views used as inputs for both algorithms are illustrated in~\figurename~\ref{fig:pattern} and the average error, in terms of PSNR, over reconstructed views is presented in~Table~\ref{table:lfpsnr}. Illustration of obtained views with corresponding difference maps is shown in~\figurename~\ref{fig:complf}. As seen from the figure, in the case of the Bunny dataset, the proposed method uses fewer views as input and still performs similar to the one in~\cite{shi2014light}. In the case of the Truck dataset, the proposed method performs significantly better in terms of average PSNR.

\begin{figure*}[th]
\centering
\includegraphics[width=7in]{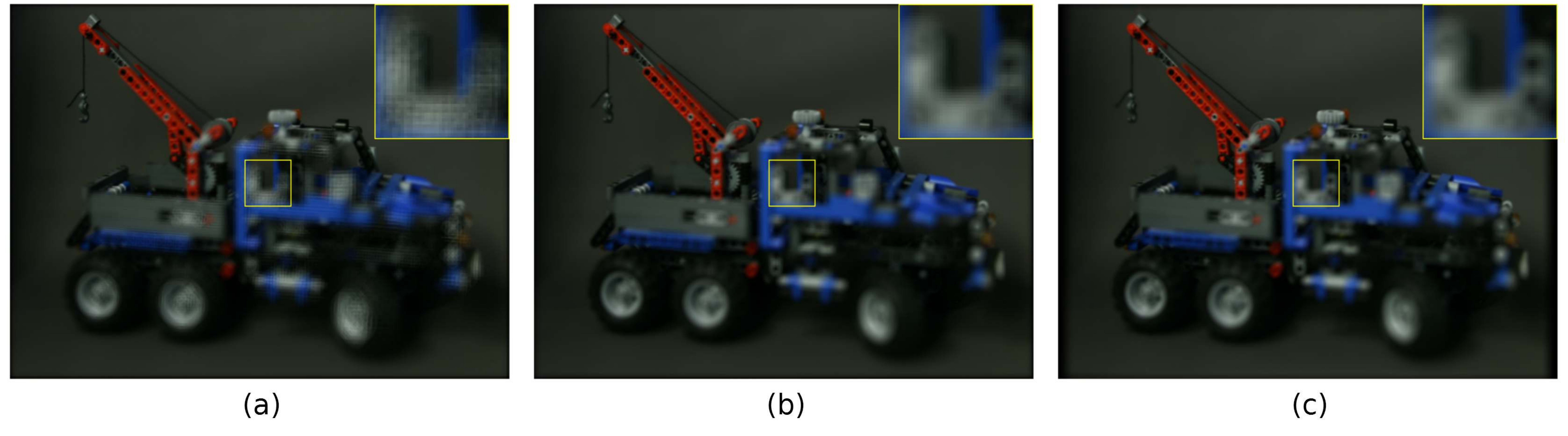}
\caption{Example of refocused images generated from differently sampled light field Truck from~\cite{vaish2008new} using linear interpolation for shearing operation. (a) Refocused image generated for central view using $5\times5$ views from original dataset, every 4th view has been chosen. (b) Refocused image generated using all $17\times17$ views from light field. (c) Refocused image generated from reconstructed light field ($49\times49$ views) based on decimated light field ($5\times5$ views).}
\label{fig:truckref}
\end{figure*}

One of the applications of full parallax LF is to construct digitally refocused images in post-processing. \figurename~\ref{fig:truckref} shows digitally refocused images corresponding to the central view for differently sampled LFs. As expected, the lack of available views results in strong artifacts in the synthesized refocused image~\figurename~\ref{fig:truckref}~(a) where only $5 \times 5$ subset of views is used, while for the up-sampled (reconstructed) LF consisting of $49 \times 49$ views, very small disparity between the reconstructed views causes smooth blurring in the refocused image areas. \figurename~\ref{fig:truckref}~(c), presents the result of similar refocusing for the original dataset~\figurename~\ref{fig:truckref}~(b).

\section{Conclusions}\label{sec:conc}
In the paper, we have presented a method for reconstructing densely sampled LF from a small number of rectified multiview images taken with a wide baseline. The reconstructed LF bears the property that the disparity between adjacent views is 1 pixel at most while the input views can be with quite high disparity. The method utilizes a sparse representation of the underlying EPIs in shearlet domain and employs an iterative regularized reconstruction. We have constructed a shearlet frame specifically for the case of EPIs and proposed an adaptive tuning for the parameter controlling the convergence in the iterative procedure. Experiments with various datasets compare our method favorably against the reference DIBR software and the state of the art in IBR. A feature of the method is that it reconstructs all views and therefore can be used in applications which require densely sampled views such as refocusing, wide field of view LF displays and digital holographic printing.

Although, the implementation of the algorithm reported in this paper is limited to scenes with Lambertian properties, it is possible to extend the algorithm such that it will be able to reconstruct non-Lambertian scenes. This will, primary, require modification of the bases used in reconstruction since different parts of the frequency domain has to be covered, in comparison to the Lambertian case. This extension is a topic of future research.

\bibliographystyle{IEEEtran}
\bibliography{bibl}

\begin{thebibliography}{10}
\providecommand{\url}[1]{#1}
\csname url@samestyle\endcsname
\providecommand{\newblock}{\relax}
\providecommand{\bibinfo}[2]{#2}
\providecommand{\BIBentrySTDinterwordspacing}{\spaceskip=0pt\relax}
\providecommand{\BIBentryALTinterwordstretchfactor}{4}
\providecommand{\BIBentryALTinterwordspacing}{\spaceskip=\fontdimen2\font plus
\BIBentryALTinterwordstretchfactor\fontdimen3\font minus
  \fontdimen4\font\relax}
\providecommand{\BIBforeignlanguage}[2]{{%
\expandafter\ifx\csname l@#1\endcsname\relax
\typeout{** WARNING: IEEEtran.bst: No hyphenation pattern has been}%
\typeout{** loaded for the language `#1'. Using the pattern for}%
\typeout{** the default language instead.}%
\else
\language=\csname l@#1\endcsname
\fi
#2}}
\providecommand{\BIBdecl}{\relax}
\BIBdecl

\bibitem{shum2008image}
S.~B. Kang, Y.~Li, X.~Tong, and H.-Y. Shum, \emph{Image-Based Rendering}.\hskip
  1em plus 0.5em minus 0.4em\relax Hanover, MA, USA: Now Publishers Inc., Jan.
  2006, vol.~2, no.~3.

\bibitem{scharstein2002taxonomy}
D.~Scharstein and R.~Szeliski, ``A taxonomy and evaluation of dense two-frame
  stereo correspondence algorithms,'' \emph{Int'l J. of Computer Vision},
  vol.~47, no. 1-3, pp. 7--42, 2002.

\bibitem{kim2013scene}
C.~Kim, H.~Zimmer, Y.~Pritch, A.~Sorkine-Hornung, and M.~Gross, ``Scene
  reconstruction from high spatio-angular resolution light fields,'' \emph{ACM
  Trans. Graph.}, vol.~32, no.~4, pp. 73:1--73:12, Jul. 2013.

\bibitem{pearson2013plenoptic}
J.~Pearson, M.~Brookes, and P.~Dragotti, ``Plenoptic layer-based modeling for
  image based rendering,'' \emph{IEEE Trans. Image Processing}, vol.~22, no.~9,
  pp. 3405--3419, Sept 2013.

\bibitem{wanner2014variational}
S.~Wanner and B.~Goldluecke, ``Variational light field analysis for disparity
  estimation and super-resolution,'' \emph{IEEE Trans. Pattern Analysis and
  Machine Intelligence}, vol.~36, no.~3, pp. 606--619, March 2014.

\bibitem{adelson1991plenoptic}
E.~H. Adelson and J.~R. Bergen, \emph{The plenoptic function and the elements
  of early vision}.\hskip 1em plus 0.5em minus 0.4em\relax Vision and Modeling
  Group, Media Laboratory, Massachusetts Institute of Technology, 1991.

\bibitem{levoy1996light}
M.~Levoy and P.~Hanrahan, ``Light field rendering,'' \emph{Proc. ACM SIGGRAPH},
  pp. 31--42, 1996.

\bibitem{gortler1996lumigraph}
S.~J. Gortler, R.~Grzeszczuk, R.~Szeliski, and M.~F. Cohen, ``The lumigraph,''
  \emph{Proc. ACM SIGGRAPH}, pp. 43--54, 1996.

\bibitem{lin2004geometric}
Z.~Lin and H.-Y. Shum, ``A geometric analysis of light field rendering,''
  \emph{Int'l J. of Computer Vision}, vol.~58, no.~2, pp. 121--138, 2004.

\bibitem{chai2000plenoptic}
J.-X. Chai, X.~Tong, S.-C. Chan, and H.-Y. Shum, ``Plenoptic sampling,''
  \emph{Proc. ACM SIGGRAPH}, pp. 307--318, 2000.

\bibitem{ng2005fourier}
R.~Ng, ``Fourier slice photography,'' \emph{ACM Trans. Graph.}, vol.~24, no.~3,
  pp. 735--744, July 2005.

\bibitem{tosic2014light}
I.~Tosic and K.~Berkner, ``Light field scale-depth space transform for dense
  depth estimation,'' \emph{Proc. IEEE Conf. Computer Vision and Pattern
  Recognition Workshops (CVPRW)}, pp. 441--448, June 2014.

\bibitem{tanimoto2006overview}
M.~Tanimoto, ``Overview of {FTV} (free-viewpoint television),'' in \emph{IEEE
  Conf. Multimedia and Expo (ICME 2009).}, June 2009, pp. 1552--1553.

\bibitem{jurik2012geometry}
J.~Jurik, T.~Burnett, M.~Klug, and P.~Debevec, ``Geometry-corrected light field
  rendering for creating a holographic stereogram,'' \emph{Proc. IEEE Conf.
  Computer Vision and Pattern Recognition Workshops (CVPRW)}, pp. 9--13, June
  2012.

\bibitem{bolles1987epipolar}
R.~Bolles, H.~Baker, and D.~Marimont, ``Epipolar-plane image analysis: An
  approach to determining structure from motion,'' \emph{Int'l J. of Computer
  Vision}, vol.~1, no.~1, pp. 7--55, 1987.

\bibitem{shi2014light}
L.~Shi, H.~Hassanieh, A.~Davis, D.~Katabi, and F.~Durand, ``Light field
  reconstruction using sparsity in the continuous fourier domain,'' \emph{ACM
  Trans. on Graphics (TOG)}, vol.~34, no.~1, p.~12, 2014.

\bibitem{hauser2012seismic}
S.~Hauser and J.~Ma, ``Seismic data reconstruction via shearlet-regularized
  directional inpainting,'' 2012.

\bibitem{vagh2015imag}
S.~Vagharshakyan, R.~Bregovic, and A.~Gotchev, ``Image based technique via
  sparse representation in shearlet domain,'' \emph{Proc. IEEE Int'l Conf. on
  Image Processing. (ICIP ’15) (accepted for publication)}, 2015.

\bibitem{liang2011light}
C.-K. Liang, Y.-C. Shih, and H.~Chen, ``Light field analysis for modeling image
  formation,'' \emph{IEEE Trans. Image Processing}, vol.~20, no.~2, pp.
  446--460, Feb 2011.

\bibitem{candes1999curvelets}
E.~J. Candes, D.~L. Donoho \emph{et~al.}, \emph{Curvelets: A surprisingly
  effective nonadaptive representation for objects with edges}.\hskip 1em plus
  0.5em minus 0.4em\relax Stanford University, 1999.

\bibitem{candes2004new}
E.~J. Cand{\`e}s and D.~L. Donoho, ``New tight frames of curvelets and optimal
  representations of objects with piecewise $c^2$ singularities,'' \emph{Comm.
  Pure Appl. Math.}, vol.~57, no.~2, pp. 219--266, 2004.

\bibitem{kutyniok2012book}
G.~Kutyniok \emph{et~al.}, \emph{Shearlets: Multiscale analysis for
  multivariate data}.\hskip 1em plus 0.5em minus 0.4em\relax Springer Science
  \& Business Media, 2012.

\bibitem{kutyniok2012shearlets}
G.~Kutyniok, J.~Lemvig, and W.-Q. Lim, ``Shearlets and optimally sparse
  approximations,'' in \emph{Shearlets: Multiscale analysis for multivariate
  data}, G.~Kutyniok and D.~Labate, Eds.\hskip 1em plus 0.5em minus 0.4em\relax
  Birkhäuser Boston, 2012, pp. 145--197.

\bibitem{donoho2001sparse}
D.~L. Donoho, ``Sparse components of images and optimal atomic
  decompositions,'' \emph{Constructive Approximation}, vol.~17, no.~3, pp.
  353--382, 2001.

\bibitem{do2005contourlet}
M.~Do and M.~Vetterli, ``The contourlet transform: an efficient directional
  multiresolution image representation,'' \emph{IEEE Trans. Image Processing},
  vol.~14, no.~12, pp. 2091--2106, Dec 2005.

\bibitem{guo2007optimally}
G.~Easley, D.~Labate, and W.-Q. Lim, ``Optimally sparse image representations
  using shearlets,'' \emph{Proc. Fortieth Asilomar Conf. Signals, Systems and
  Computers (ACSSC '06)}, pp. 974--978, Oct 2006.

\bibitem{kutyniok2011compactly}
G.~Kutyniok and W.-Q. Lim, ``Compactly supported shearlets are optimally
  sparse,'' \emph{J. of Approximation Theory}, vol. 163, no.~11, pp. 1564 --
  1589, 2011.

\bibitem{lim2013nonseparable}
W.-Q. Lim, ``Nonseparable shearlet transform,'' \emph{IEEE Trans. Image
  Processing}, vol.~22, no.~5, pp. 2056--2065, May 2013.

\bibitem{kutyniok2014shearlab}
G.~Kutyniok, W.-Q. Lim, and R.~Reisenhofer, ``{ShearLab} {3D}: Faithful digital
  shearlet transforms based on compactly supported shearlets,'' \emph{{ACM}
  Trans. on Mathematical Software}, vol.~42, no.~1, 2015.

\bibitem{mallat2008wavelet}
S.~Mallat, \emph{A Wavelet Tour of Signal Processing : The Sparse Way},
  3rd~ed.\hskip 1em plus 0.5em minus 0.4em\relax Academic Press, 2008.

\bibitem{starck2005morphological}
J.-L. Starck, Y.~Moudden, J.~Bobin, M.~Elad, and D.~L. Donoho, ``Morphological
  component analysis,'' \emph{Proc. SPIE Optics \& Photonics}, vol. 5914, pp.
  59\,140Q--59\,140Q--15, 2005.

\bibitem{fadili2009mcalab}
J.~Fadili, J.-L. Starck, M.~Elad, and D.~Donoho, ``Mcalab: Reproducible
  research in signal and image decomposition and inpainting,'' \emph{Computing
  in Science Engineering}, vol.~12, no.~1, pp. 44--63, Jan 2010.

\bibitem{blumensath2010normalized}
T.~Blumensath and M.~Davies, ``Normalized iterative hard thresholding:
  Guaranteed stability and performance,'' \emph{IEEE J. Sel. Topics Signal
  Processing}, vol.~4, no.~2, pp. 298--309, April 2010.

\bibitem{hauser2012fast}
S.~H{\"a}user and G.~Steidl, ``Fast finite shearlet transform,'' \emph{arXiv
  preprint arXiv:1202.1773}, 2014.

\bibitem{tanimoto2009depth}
M.~Tanimoto, T.~Fujii, K.~Suzuki, N.~Fukushima, and Y.~Mori, ``Depth estimation
  reference software (ders) 5.0,'' \emph{ISO/IEC JTC1/SC29/WG11 M}, vol. 16923,
  2009.

\bibitem{tanimoto2009view}
M.~Tanimoto, T.~Fujii, and K.~Suzuki, ``View synthesis algorithm in view
  synthesis reference software 2.0 (vsrs2.0),'' \emph{ISO/IEC JTC1/SC29/WG11
  M}, vol. 16090, 2009.

\bibitem{kovacs2015bbb}
P.~Kovacs, A.~Fekete, K.~Lackner, V.~Adhikarla, A.~Zare, and T.~Balogh, ``Big
  buck bunny light-field test sequences,'' \emph{International Organisation For
  Standardisation, MPEG contribution (ISO/IEC, JTC1/SC29/WG11 M35721)},
  February 2015.

\bibitem{ToyohiroNagoya}
S.~Toyohiro, ``Nagoya university multi-view sequences,''
  \emph{http://www.fujii.nuee.nagoya-u.ac.jp/multiview-data}.

\bibitem{scharstein2003high}
D.~Scharstein and R.~Szeliski, ``High-accuracy stereo depth maps using
  structured light,'' \emph{Proc. IEEE Conf. Computer Vision and Pattern
  Recognition (CVPR)}, vol.~1, pp. I--195--I--202, June 2003.

\bibitem{vaish2008new}
V.~Vaish and A.~Adams, ``The (new) stanford light field archive,''
  \emph{http://lightfield.stanford.edu}, 2008.

\end{thebibliography}

\begin{IEEEbiography}{Suren Vagharshakyan}
Biography text here.
\end{IEEEbiography}

\begin{IEEEbiographynophoto}{Robert Bregovic}
Biography text here.
\end{IEEEbiographynophoto}

\begin{IEEEbiographynophoto}{Atanas Gotchev}
Biography text here.
\end{IEEEbiographynophoto}

\end{document}